\documentclass{article}

\usepackage{microtype}
\usepackage{graphicx}
\usepackage{subfigure}
\usepackage{booktabs} 
\usepackage{multicol}
\usepackage[linesnumbered,ruled,vlined]{algorithm2e}
\usepackage{caption}
\usepackage{pifont}
\usepackage{listings}

\usepackage[hyphens]{url}
\usepackage{color}
\definecolor{lightgray}{rgb}{.9,.9,.9}
\definecolor{red}{rgb}{.9,.1,.1}
\definecolor{darkgray}{rgb}{.4,.4,.4}
\definecolor{purple}{rgb}{0.65, 0.12, 0.82}

\lstdefinelanguage{JavaScript}{
  keywords={typeof, new, true, false, catch, function, return, null, catch, switch, var, if, in, while, do, else, case, break},
  keywordstyle=\color{blue}\bfseries,
  ndkeywords={class, export, boolean, throw, implements, import, this},
  ndkeywordstyle=\color{darkgray}\bfseries,
  identifierstyle=\color{black},
  sensitive=false,
  comment=[l]{//},
  morecomment=[s]{/*}{*/},
  commentstyle=\color{purple}\ttfamily,
  stringstyle=\color{red}\ttfamily,
  morestring=[b]',
  morestring=[b]"
}

\usepackage{hyperref}

\usepackage[accepted]{sysml2019}

\begin{document}

\twocolumn[
\sysmltitle{Fabrik: An online collaborative neural network editor}



\sysmlsetsymbol{equal}{*}

\begin{sysmlauthorlist}
\sysmlauthor{Utsav Garg}{NTU}
\sysmlauthor{Viraj Prabhu}{GT}
\sysmlauthor{Deshraj Yadav}{GT}
\sysmlauthor{Ram Ramrakhya}{INMOBI}
\sysmlauthor{Harsh Agrawal}{GT}
\sysmlauthor{Dhruv Batra}{GT,FAIR} \\
\texttt{{\href{fabrik.cloudcv.org}{fabrik.cloudcv.org}}}
\end{sysmlauthorlist}

\sysmlaffiliation{NTU}{Nanyang Technological University}
\sysmlaffiliation{GT}{Georgia Institute of Technology}
\sysmlaffiliation{INMOBI}{Inmobi}
\sysmlaffiliation{FAIR}{Facebook AI Research}

\sysmlcorrespondingauthor{Utsav Garg}{gargutsav96@gmail.com}
\sysmlcorrespondingauthor{Viraj Prabhu}{virajp@gatech.edu}

\sysmlkeywords{Machine Learning, SysML}

\vskip 0.3in

\begin{abstract}
We present Fabrik, an online neural network editor that provides tools to visualize, edit, and share neural networks from within a browser. Fabrik provides a simple and intuitive GUI to import neural networks written in popular deep learning frameworks such as Caffe, Keras, and TensorFlow, and allows users to interact with, build, and edit models via simple drag and drop. Fabrik is designed to be framework agnostic and support high interoperability, and can be used to export models back to any supported framework. Finally, it provides powerful collaborative features to enable users to iterate over model design remotely and at scale.
\end{abstract}
]



\printAffiliationsAndNotice{}  

\section{Introduction}
\label{introduction}


\par \noindent
In recent years, long strides have been made in artificial intelligence, primarily propelled by the advent of deep learning~\cite{lecun2015deep}. Deep neural networks have been successfully applied to a range of diverse applications~\cite{krizhevsky2012imagenet,bahdanau2014neural, ren2015faster, antol2015vqa, visdial, embodiedqa, silver2016mastering}, and several other avenues remain to be explored. A catalyst to this progress has been the development of well-designed, well-documented, open source frameworks for deep learning~\cite{abadi2016tensorflow,jia2014caffe,paszke2017pytorch,chollet2017keras}. 

\par \noindent
There has been an explosion in the number of such deep learning frameworks -- over 23 have been released as of last year~\cite{kdnuggets2017survey}. Several of these are optimized for specific use cases, and have distinct strengths, such as efficient deployment and serving~\cite{abadi2016tensorflow}, efficient inference on mobile devices ~\cite{caffe2ai2018}, or easy extensibility for rapid research and prototyping~\cite{paszke2017pytorch}. However, keeping up with the latest frameworks presents a daunting challenge to newcomers and experienced researchers alike. To the large number of newcomers trying to enter the field, the need to become familiar with multiple frameworks presents a high barrier to entry. For instance, code for a state of the art paper they read may be published in a new framework they are completely unfamiliar with. Experienced researchers as well tend to have a preferred framework in which they have developed expertise. Having to constantly learn new frameworks (say, to run a model written using an unfamiliar framework, perhaps as a baseline), regularly proves to be a significant time sink.
\par \noindent
At the same time, neural networks developed in such frameworks are often designed \textit{collaboratively} over the course of several iterations. However, the process of this model design and development is still largely restricted to whiteboard discussions, and scaling such collaboration to remote and online settings remains an open problem. In recent years, there have been many valuable efforts in developing tools for networked science~\cite{nielsen2011reinventing}. Such tools greatly aid scientists with complementary expertise to collaborate more efficiently~\cite{vanschoren2014openml}. Considering the popularity of deep learning, a tool for collaborative design of neural networks is likely to have widespread utility.
{\setlength\intextsep{0pt}
\begin{figure}[t]
    \centering
    \includegraphics[width=\linewidth]{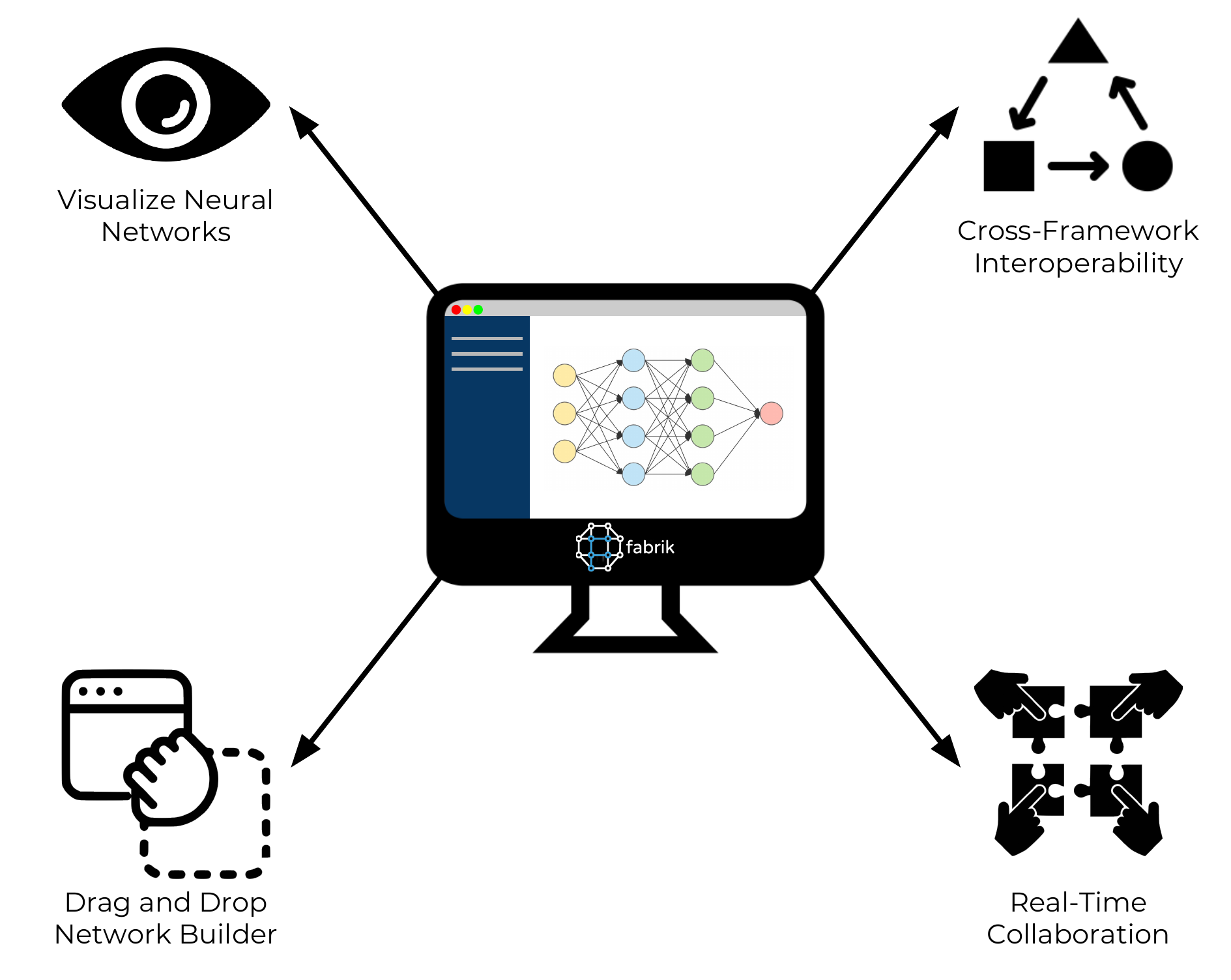}
    \vspace{-20pt}
    \caption{We present Fabrik, an online collaborative neural network editor to visualize, edit, and share networks.}
    \label{fig:teaser}
    \vspace{-10pt}
\end{figure}
}

\par \noindent 
As an attempt to solve these challenges, we present Fabrik, an online collaborative neural network editor. Fabrik is an open source platform providing tools to visualize, edit, and share neural networks from within the browser. See Figure~\ref{fig:teaser}.
\vspace{-20pt}
\begin{itemize}
    \item \textbf{Visualize.} Fabrik provides a simple and intuitive GUI to import neural networks written in popular deep learning frameworks such as Caffe, Keras, and TensorFlow.\footnote{Note that these are the top three frameworks by popularity based on Github metrics~\cite{zacharias2018survey}.}  Further, it also supports building a neural network via a simple drag and drop interface.
    \item \textbf{Edit.} Fabrik supports editing layers and parameters of the neural network right in the browser. Once completed, a network can easily be exported to a target framework, which can be different from the one from which it is imported. Fabrik is designed to be framework agnostic and provide easy interoperability between different frameworks.
    \item {\textbf{Share.}} Fabrik provides powerful collaborative features for researchers to brainstorm and design models remotely. A researcher can generate a shareable link for a model, and can see real time updates and comments from their collaborators. Such a link can be published along with deep learning research publications to provide an aesthetic and interactive visualization of the model proposed, enabling reproducibile research.
\end{itemize}
\par\noindent
Fabrik is targeted at newcomers and experienced deep learning practitioners alike. For newcomers, it lowers the barrier to entry to the field and enables them to interact with neural networks in the browser without having to grapple with syntax. We also think of Fabrik as `science as a service' -- an accessible teaching tool for newcomers to learn the nuts and bolts of deep learning `by doing'.

\par \noindent
For researchers, it provides a tool to easily visualize the depth and breadth of their network and verify correctness of implementation. It further supports porting between frameworks to enable that they can continue to work in their preferred framework. Finally, it provides collaborative features to brainstorm about model design remotely and at scale, and to share interactive visualizations of their models to aid in reproducibility.
\par \noindent
In the following sections we describe our application in detail, and present our approach to generating visualizations and supporting interoperability between multiple deep learning frameworks. We also walk through concrete use cases of this tool that illustrate its various functionalities.

\section{Related Work}
\label{relWorks}

\textbf{Deep learning in the browser.} A number of browser-based deep learning tools have been released recently. Some of these are geared towards provide an online platform to completely manage and run deep learning experiments~\cite{Aetros, KerasJS}. However, these are either not free~\cite{Aetros}, or support only a single framework~\cite{KerasJS}. Yet others, like \citet{ConvNetJS} have been designed primarily as teaching tools and do not plug into existing popular frameworks. In this work we propose Fabrik as a tool for easy visualization, export, and import of models between a range of popular deep learning frameworks.

\textbf{Visualizing Neural Networks.} A few tools specifically designed for generating neural network visualizations have been proposed recently. For example, \citet{Netscope} supports visualizing arbitrary directed acyclic graphs represented in Caffe's prototxt format. However, it works only with Caffe models, and only solves a part of the problem by supporting import and visualizations of existing models. \citet{Tensorboard} is a web application designed for inspecting and debugging models written in TensorFlow, which also supports visualizing computation graphs. While a powerful tool, it requires writing code to set up, and is thus not as accessible to newcomers. Moreover, it only works with TensorFlow and related frameworks such as Keras. Fabrik supports interactively building models from within the browser, collaborating with other researchers, and exporting to popular frameworks, without having to write a line of code.

\textbf{Cross-framework support.} A few tools have recently been proposed that promote interoperability between deep learning frameworks~\cite{ONNX, MMdnn}. \citet{ONNX} provides an open source format for AI models, and is supported by frameworks such as Caffe2, PyTorch, Microsoft Cognitive Toolkit, and Apache MXNet. A notable exception in this list is TensorFlow, which is presently the most popular deep learning framework. We view ONNX as an alternative intermediate format that be integrated into our application in the future for long-term maintainability. Similarly, \citet{MMdnn} is a cross-framework solution to convert between and visualize deep neural network models written in several popular frameworks. However, it does not provide tools for building models from within the browser. Additionally, it is not an online service and is not designed for remote collaboration.


\section{Application Overview}
\label{application}


Fabrik can be represented by a 3-tier architecture, with the tiers being broadly classified as Presentation, Intermediate Representation, and Back-End. The following subsections discuss each of these components in detail. 

\subsection{Presentation Layer}
\begin{figure*}[ht]
    \centering
    \includegraphics[width=0.9\linewidth]{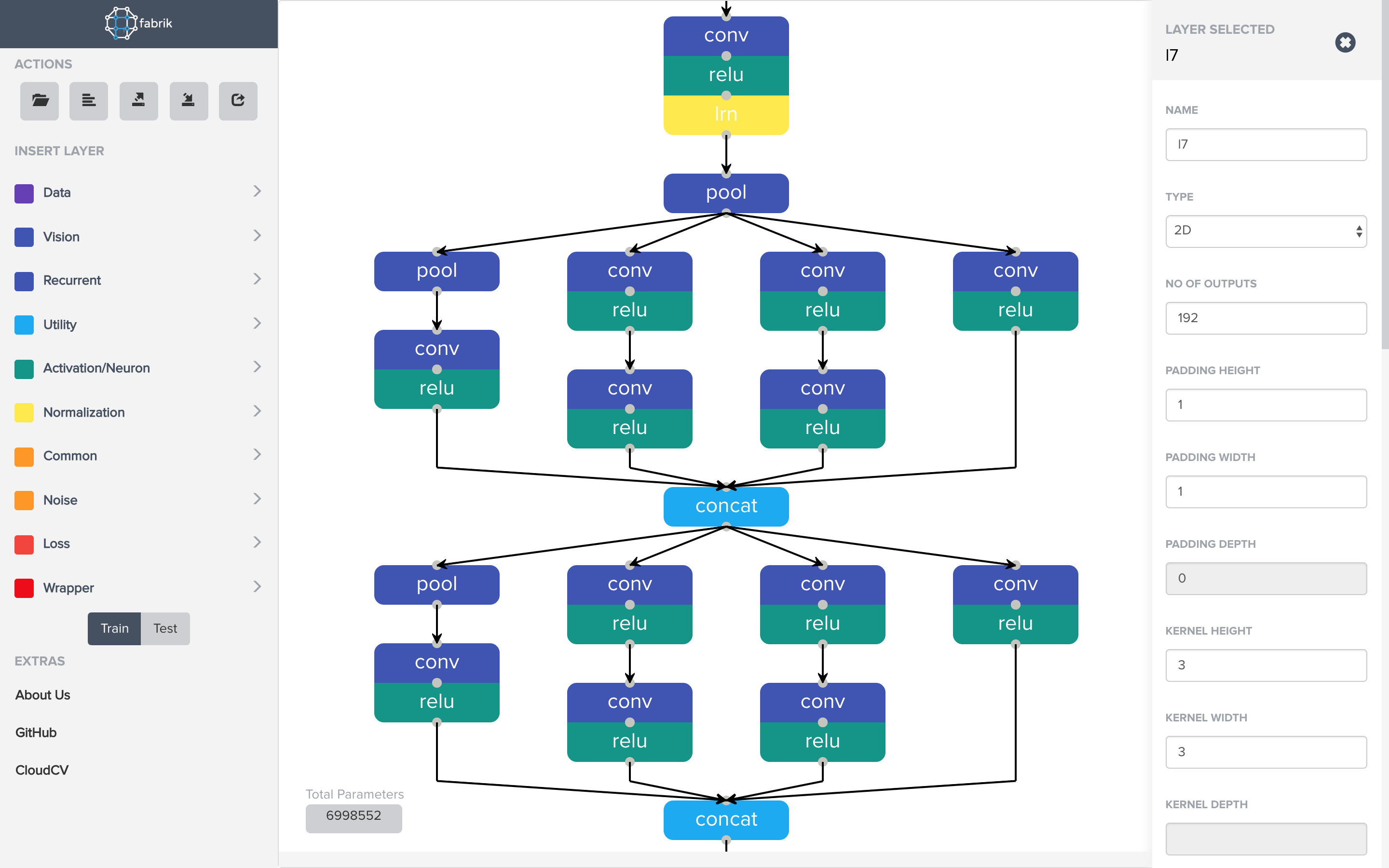}
    \caption{Visualizing the GoogLeNet~\cite{szegedy2015going} architecture with Fabrik.}
    \label{fig:overview}
\end{figure*}

The user-facing Presentation layer is one of the most important components as all interactions with the user depend on it. Figure~\ref{fig:overview} shows the landing screen displayed with the GoogLeNet~\cite{szegedy2015going} network being visualized.

\begin{figure}[h!]
    \centering
    \includegraphics[width=0.7\linewidth]{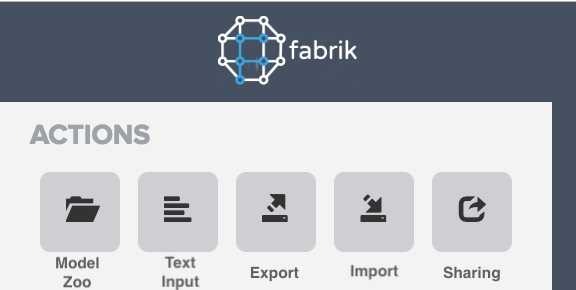}
    \caption{The main dashboard menu provides options to import, export, and share models.}
    \label{fig:topMenu}
\end{figure}

We can see there are two distinct sections -- on the right is the \emph{Canvas} where the neural network is displayed and on the left is the \emph{Dashboard}, which provides all of Fabrik's core functionalities, such as adding layers, importing or exporting models, or sharing models with collaborators.

\begin{figure*}[t]
    \centering
    \begin{minipage}{.48\textwidth}
        \centering
        \includegraphics[height=8cm]{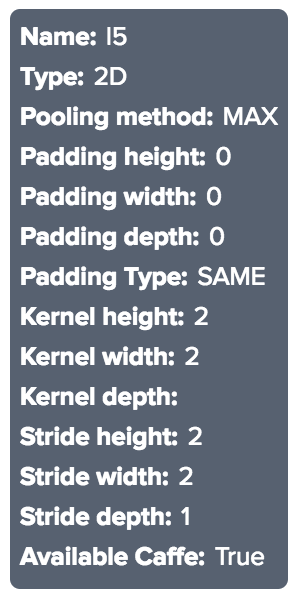}
        \caption{A tool-tip is rendered on hover, providing a quick overview of layer parameters, in this case a 2D convolutional layer.}
        \label{fig:layerHover}
    \end{minipage}\hfill
    \begin{minipage}{.48\textwidth}
        \centering
        \includegraphics[height=8cm]{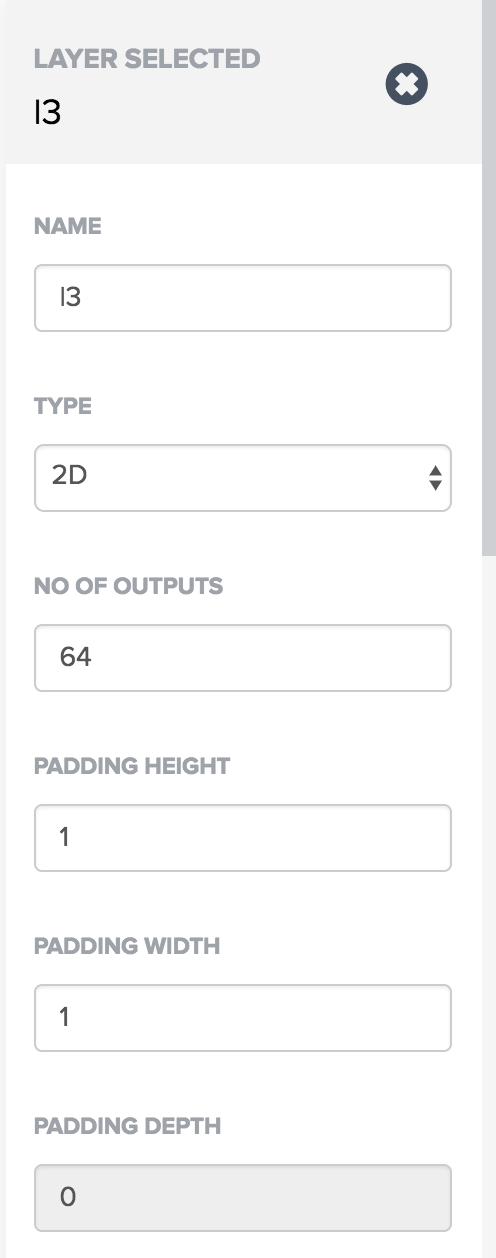}
        \caption{A side panel is displayed when a layer is clicked on to allow edits to layer parameters. Note that in this case padding depth has been grayed out as unavailable, because the layer is two dimensional.}
        \label{fig:layerClick}
    \end{minipage}
\label{fig:canvasFeatures}
\end{figure*}

\subsubsection{Canvas}
The canvas displays the neural network being visualized. 
The network is visualized using the jsPlumb~\cite{jsPlumb} library that allows for a simple interface for connecting UI elements. The layers of the network are represented as nodes of a graph which are connected based on layer dependencies. These nodes can be dragged and re-positioned on the canvas. Upon hovering over a layer, an overview of all layer parameters is displayed as a tool-tip, as demonstrated in Figure~\ref{fig:layerHover}. 
When a layer in the network is hovered over, a succinct overview of all layer parameters is displayed. 
On clicking a layer, a side-pane opens up where its parameters can be updated, as shown in Figure~\ref{fig:layerClick}. In cases where the layer parameters might depend on the dimensionality of the input (for example, 1D, 2D and 3D convolutions), only applicable parameters are available to be edited by the user.
Notice how the depth dimension has been blocked in this case as the layer is two dimensional. 

\subsubsection{Dashboard}
The dashboard provides controls for all core functionalities -- export, import, building, and sharing. Figure~\ref{fig:topMenu} shows the top menu options of this sidebar. The first button accesses the Model Zoo, a collection of state-of-art deep learning models that we provide inbuilt support for (Table~\ref{tab:testedModels} provides a complete list). These models can be directly loaded into the application and serve as samples for users to work with. The next three buttons (labeled as Text Input, Export and Import in Figure~\ref{fig:topMenu}) allow the user to initiate import/export of their neural network model. Under each of these options, the user is prompted to choose the source or target framework for which they wish to perform this operation. Finally, the last button allows sharing the current model with other users through a unique generated URL.

\begin{figure*}[th]
    \centering
    \begin{minipage}{.48\linewidth}
        \centering
        \includegraphics[width=0.7\linewidth]{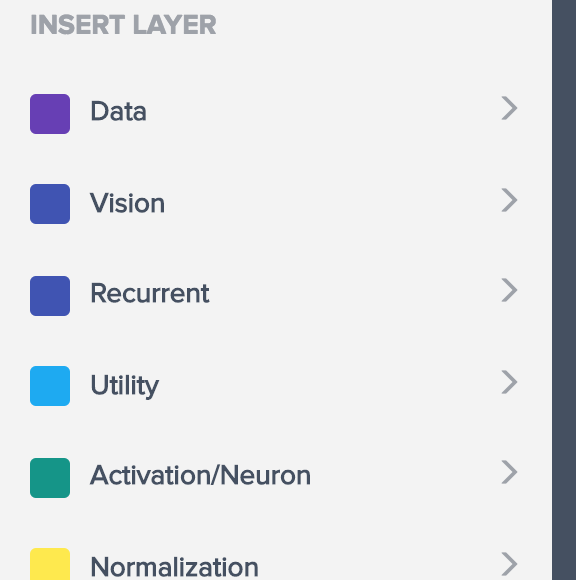}
        \caption{The layer categories available in the sidebar, which can be expanded to reveal a complete list of layers.}
        \label{fig:layersCollapsed}
    \end{minipage}
    \hfill
    \begin{minipage}{.48\linewidth}
        \centering
        \includegraphics[width=0.7\linewidth]{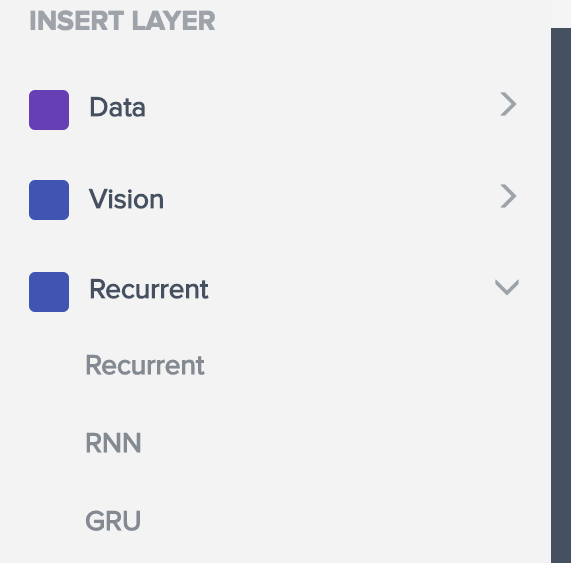}
        \caption{Expanded view of one of the Recurrent catgory. Any of the Recurrent, RNN, and GRU layers can now be dragged and dropped onto the canvas as a new layer.}
        \label{fig:layerExpanded}
    \end{minipage}
\end{figure*}

 Below the top menu are layers that can be dragged and dropped onto the canvas to build or edit a model. Figure~\ref{fig:layersCollapsed} shows the category of layers available. Each top level category drills down to a set of layers, as shown in Figure~\ref{fig:layerExpanded}. These layers can either be dragged onto the canvas, connected as desired by the user, or can simply be clicked, upon which the application will automatically connect it to the deepest layer in the current network being visualized. On adding a layer, default parameter values are provided which can be updated by the user as required.

\subsubsection{Visualization Algorithm}
The visualization algorithm plots the layer objects on the canvas. The main goal of the algorithm is to assign coordinates and establish connections between layer objects to yield an accurate and aesthetic visualization of the model. Each node in the graph is a layer of the deep learning model and a connection between layer objects represents layer dependencies. The algorithm focuses on minimizing the number of overlapping layers and connections in order to achieve a clean visualization. A few challenges that our visualization algorithm addresses include:

\begin{algorithm}[h!]
  \caption{Visualization Algorithm for Fabrik}
  \label{alg:visualization}
  \SetKwRepeat{Search}{search}{}
    \KwIn{modelGraph}
    \KwOut{positionMap}
    
    \ForEach{node \textbf{in} modelGraph}{compute $inDegree$ \& $outDegree$}
    \textbf{create} stack S \& push source node with indegree 0
    
    \While{not S.empty()}{
        $layerId \leftarrow S.pop()$
        
        \textbf{compute} $parentLayerId$ \& $inputLength$
        
        \If{$parentLayerId$}{
            $outputLength \leftarrow outDegree[parentLayerId]$
        }
        
        \uIf{not $parentLayerId$}{
            $positionMap[layerId] \leftarrow [0,0]$
        }
        \uElseIf{$inputLength=1$ \& $outputLength=1$}{
            allocate position to center
        }
        \uElseIf{$inputLength>1$}{
            allocate position to current layer at max depth in the center
        }
        \ElseIf{$inputLength=1$ \& $outputLength\neq1$}{
            allocate positions to aligning child layers horizontally
        }
    
    \ForEach{childLayerId \textbf{in} modelGraph[layerId].output}{
        $layerIndegree[childLayerId]$ -= 1
        
        \If{$layerIndegree[childLayerId]=0$}{
            $S.push(childLayerId)$
        }
    }
    
    \textbf{mark} $layerId$ as visited
    
    \While{layer overlap on left or right}{
            increment Y-coord
        }
    
    Update layer position with X and Y coord
    }
\Return{positionMap}
\end{algorithm}
\begin{enumerate}
    \item \textbf{Overlapping layers:} In order to obtain a good visualization, the algorithm should be able to allocate positions to layers so as to avoid or minimize overlap. To achieve a non-overlapping position allocation, we implement a hashing technique: maintaining a constant height and width for each layer object, an initial position allocation is computed. Each time a layer has been assigned co-ordinates, the immediate area surrounding the layer is searched with the help of hashed co-ordinates of visited layers to detect overlap. If overlap is detected, a constant displacement is added to the y-coordinate of the layer, and we then check for overlap recursively. The search space for detecting overlap is reduced significantly via hashing, leading to much faster visualization. 
    \item \textbf{Order of traversal:} Topological sorting in depth-first order is used for the purpose of assigning non-overlapping positions. This ensures that a layer object is visited only after all of its parent layers have been assigned positions. With the help of the topological depth-first order of traversal we reduce the search space for overlapping layers to only six directions i.e. top-left, top-right, bottom-left, bottom-right, left, and right. Using the algorithm we assign layer positions from top to bottom and right to left. All sibling layers fall in the same horizontal space and all the child layers fall in the same vertical space. The algorithm also handles corner cases like connection cycles in a model graph via an efficient implementation of cycle detection.
\end{enumerate}

Once layers have been assigned non-overlapping positions, connections between layer nodes are built with the help of \citet{jsPlumb}. To ensure non-overlapping connections, various parameters are considered, e.g. the number of child nodes of a layer, position of the source layer, and position of the destination layer. Based on these parameters, the slope of the connection arrow is calculated, and if needed the line is divided into sub parts to ensure that connection overlap and cutting connections (i.e. connections which cut through a layer object) are minimized. Algorithm~\ref{alg:visualization} describes the steps of our visualization process.

Several parameters for this rendering pipeline are pre-computed in order to improve the efficiency of algorithm. In order to optimize the depth first topological traversal pre-computation of in-degree (i.e. the number of incoming connections) and out-degree (i.e. number of outgoing connections) of every layer in the model graph allows us to reduce the complexity from $O(|V|^2)$ to $O(|V| + |E|)$, where $|V|$ is the number of layers and $|E|$ is the number of connections in model graph. In addition to in-degree and out-degree, pre-computing input-length (i.e. the number of parent layers) and output-length (i.e. the number of child layers) improves the efficiency of the allocation logic.

\subsection{Intermediate Representation Layer}
\label{ref:intermediate}
The intermediate representation allows us to decouple the display portion of the application from framework specific back-ends. On importing a model from any of the supported frameworks, the layers are stored as the corresponding layers in the intermediate representation. We use the prototxt-based model configuration protocol employed by Caffe~\cite{jia2014caffe} as our intermediate representation as it is clean and extensible. Listing~\ref{lst:acc} shows an example of the intermediate representation for the `accuracy' layer. The comment at the top indicates that this layer is only available for the Caffe framework. This is further monitored by the 'caffe' parameter. The parameters are used to specify layer properties and the inputs are restricted by certain conditions. For example, the \emph{$top\_k$} field tells the framework the top-k predictions of the network are to be considered, and that the field is restricted to be numerical.

\begin{lstlisting}[caption=Intermediate Representation for the Accuracy layer, label={lst:acc}]
  Accuracy: { // Only Caffe
    name: 'acc',
    color: '#f44336',
    endpoint: {
      src: [],
      trg: ['Top']
    },
    params: {
      top_k: {
        name: 'Top-K',
        value: 1,
        type: 'number',
        required: false
      },
      axis: {
        name: 'Axis',
        value: 1,
        type: 'number',
        required: false
      },
      caffe: {
        name: 'Available Caffe',
        value: true,
        type: 'checkbox',
        required: false
      }},
    props: {
      name: {
        name: 'Name',
        value: '',
        type: 'text'
      }},
    learn: false
  }
\end{lstlisting}
As different frameworks might have slightly varying parameters or parameter names, we first map them to corresponding names in this representation. Note that we extend the protocol to hold the union of all parameters required by all supported frameworks, so that no information is lost. 

Creating an intermediate representation that serves all supported frameworks well is a challenging task. For example, the Caffe framework uses numerical padding, to pad the inputs before applying operations such as Convolution or Pooling. But frameworks like Keras and Tensorflow use 'SAME' or 'VALID' padding, where the amount of padding required is automatically determined based on input and output sizes. To map from both types of frameworks to the same intermediate representation, we first need to convert the 'SAME'/'VALID' type of padding to its numerical counterpart by determining the input and output sizes for each layer, and perform the reverse operation when converting back. Note that while we provide a single example here, many such corner cases exist and are apporpriately handled by the back-end layer.

\subsection{Back-End Layer}
The back-end layer implements Fabrik's business logic.

\subsubsection{Model conversion}
In addition to helping us decouple the visualization from framework specific layers, the intermediate representation helps support conversion from one framework to another. As different frameworks might have different naming conventions, a name mapping is used for each framework. Without such an intermediate representation, separate logic would be required to port between each pair of supported frameworks, but now the export logic is only required to be written from the intermediate representation to each of the supported framework. This reduces the problem complexity significantly from $O(n^2)$ to $O(n)$, for $n$ 
different frameworks. 


In some cases, model conversion is restricted by major differences between frameworks, such as certain layers being exclusive to a framework. For example, the \emph{Local Response Normalization (LRN)} layer is only available in Caffe, and so it is not natively possible to port an AlexNet~\cite{krizhevsky2012imagenet} network defined in Caffe (which uses this layer) to Keras or Tensorflow. In some cases where native support is missing, we complete the conversion by using third-party implementations. For \emph{LRN}, we use such implementations to allow export to these frameworks. However, there still remain cases where even this is not possible. For example, Caffe supports a \emph{Python} layer that allows users to implement any custom logic. This can clearly not be ported to other frameworks, and in this case we report an appropriate error message.
Note that since TensorFlow is a supported backend of Keras, we actually need to define porting logic for just two frameworks, Caffe and Keras. For TensorFlow conversion, we first export the model as Keras and then internally use the Keras backend to port to TensorFlow. 

\subsubsection{Asynchronous tasks for Model Export}
\label{async}
For a model to be exported to a target framework, a back-end task need to be executed which handles the parsing of the intermediate model representation. The back-end task for model export works by converting the intermediate representation to a model object of the respective framework, which is then used to generate a saved graph of the model architecture. Waiting for the task to complete might prevent a user from being able to perform any operations on the canvas. To overcome this, asynchronous tasks for model export are employed.

\begin{figure}[th]
\centering
    \includegraphics[width=\linewidth]{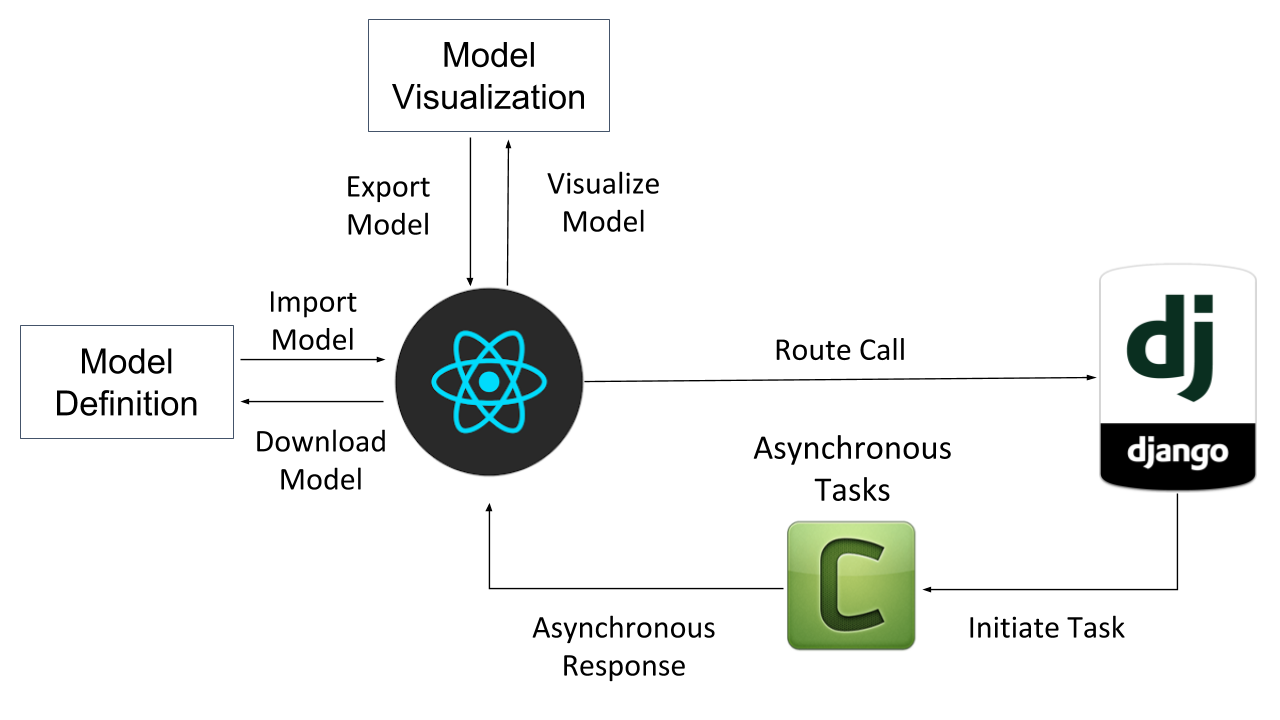}
    \caption{Export / Import Workflow}
    \label{fig:exportWorkflow}
\end{figure}

The asynchronous tasks are implemented using Celery~\cite{Celery}, which allows for distributing work across threads or machines. {Figure~\ref{fig:exportWorkflow}} shows how celery workers are used in the Fabrik infrastructure. Whenever a model export event is triggered, the intermediate representation preprocessing is completed at the front-end, following which this intermediate representation is passed on to the back-end. Based on the type of framework for which the export event is triggered, the respective celery task is executed. As the call to the export event is implemented in an asynchronous manner, there is no latency at the front-end and the user can continue their operations. Meanwhile at the back-end, the export request is allocated depending upon the availability of Celery workers, which completes the generation of the respective export file for the model graph. Once the task completes it sends a response containing the result of the task with the help of web-sockets. 
Finally, once the front-end layer receives a response message for model export, the user is notified with the result of the task. By using asynchronous tasks for model export we ensure that users are not stalled due to a time consuming back-end functionality.

\subsubsection{Real-time collaboration}
Fabrik provides support for collaboratively building deep learning models with multiple users working in real time on a single model. For sharing a model with multiple users, a unique link for a model is generated and any user with access to the open link can view, edit, or comment on a model. A global copy of a single shared model is maintained in the database and all updates over multiple sessions are performed on this copy, using communication via web-sockets. Whenever an update operation is performed, an event message is broadcast by the back-end layer which is used by presentation layer to rerender the canvas on all active sessions. Additionally, the layer being updated is highlighted with the name of the user performing the update.
     
In order to achieve real-time collaboration, we employ an efficient implementation of operational transformation for the intermediate representation of a model. By storing the intermediate representation of a model in JSON format, the size of this update is minimized. Based on the type of operation being performed, there are 4 types of operational update events -- parameter update, layer addition, layer deletion, and layer highlight. Whenever one of these is performed, a corresponding update event is triggered and propagated to all active sessions for maintaining consistency in the collaborative session.

\section{Deployment}
\begin{figure*}[th]
\centering
\includegraphics[width=0.7\linewidth]{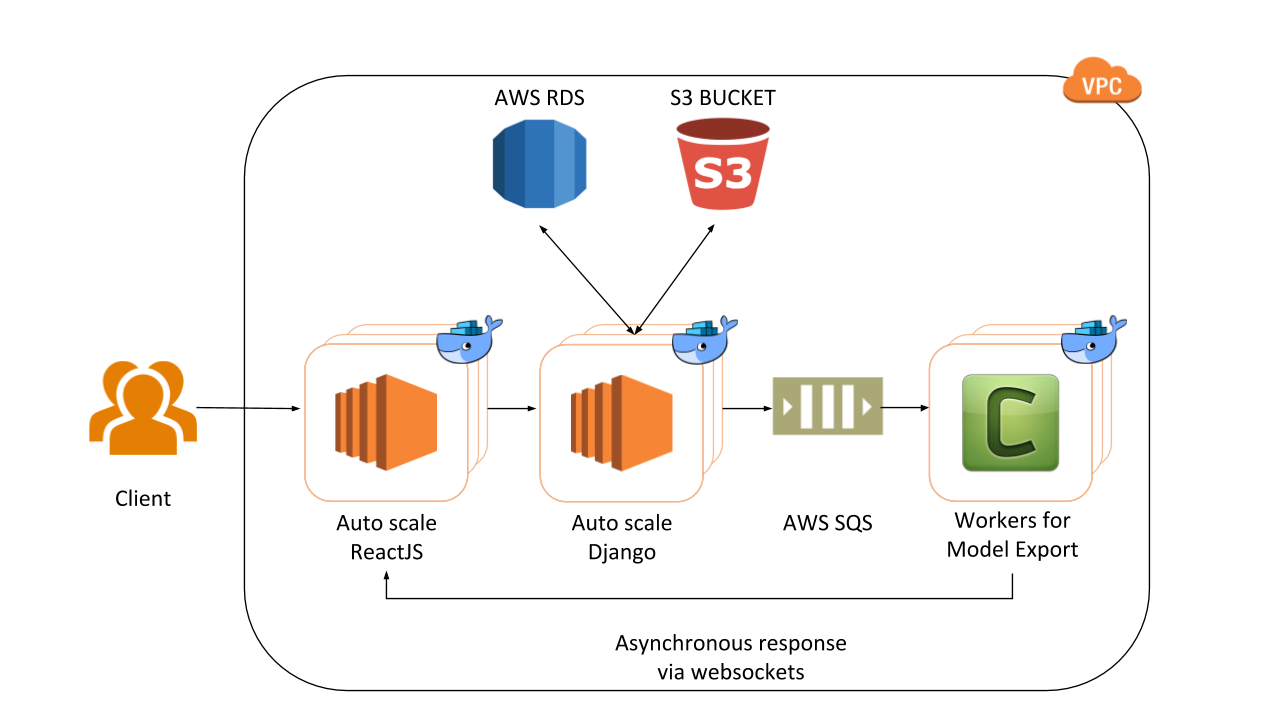}
\caption{Fabrik System architecture}
\label{fig:systemArchitecture}
\end{figure*}

In recent years, containerized deployment has received widespread adoption from the software engineering community. We use a Docker based setup which allows us to easily pack, distribute, and manage applications within containers, and makes scaling easy. Figure~\ref{fig:systemArchitecture} demonstrates our system architecture. Currently, we have five services that serve as the backbone of Fabrik:
        \textbf{ReactJS.} We use ReactJS to build our frontend. All user requests go to the backend through ReactJS as REST API calls.
        \textbf{Django.} We use Django,a high-level python based MVC Framework for our backend. It is responsible for serving REST APIs and import and export route call controlling. 
        \textbf{PostgreSQL.} We use PostgreSQL, an object-relational database management system, to store user and model import and export related data.
        \textbf{Redis.} Redis is a fast in-memory database that is mainly used to power real-time internet applications. We use Redis to support our  real-time collaborative environment, to store session data, and support low-latency fetching.
        \textbf{Celery.} As described in ~\ref{async}, we use ~\citet{Celery} for processing heavy jobs (such as exporting GoogLenet) at the backend without blocking I/O for other operations, which runs these operations asynchronously.

\begin{figure*}[th!]
    \centering
    \begin{minipage}[b]{.22\linewidth}
        \centering
        \includegraphics[width=\linewidth]{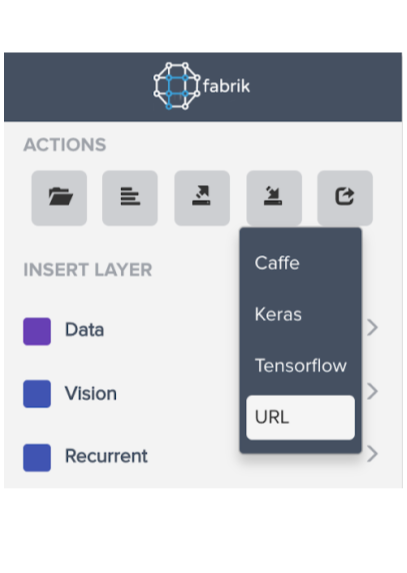}
        (a)
    \end{minipage}\hfill
    \begin{minipage}[b]{.3\linewidth}
        \centering
        \includegraphics[width=\linewidth]{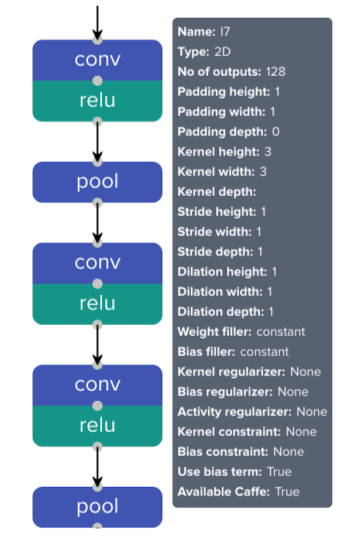}
        (b)
    \end{minipage}\hfill
    \begin{minipage}[b]{.2\linewidth}
        \centering
        \includegraphics[width=\linewidth]{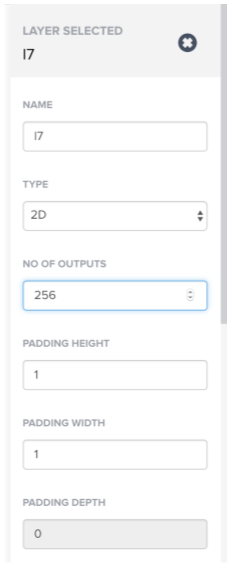}
        (c)
    \end{minipage}\hfill
    \begin{minipage}[b]{.22\linewidth}
        \centering
        \includegraphics[width=\linewidth]{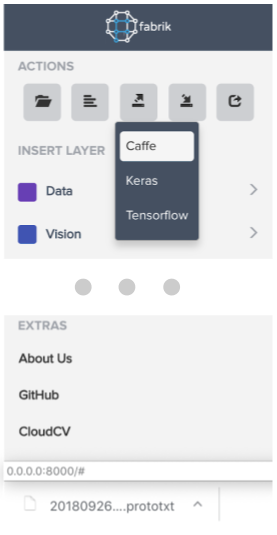}
        (d)
    \end{minipage}
    \begin{minipage}{.75\linewidth}
        \centering
        \includegraphics[width=\linewidth]{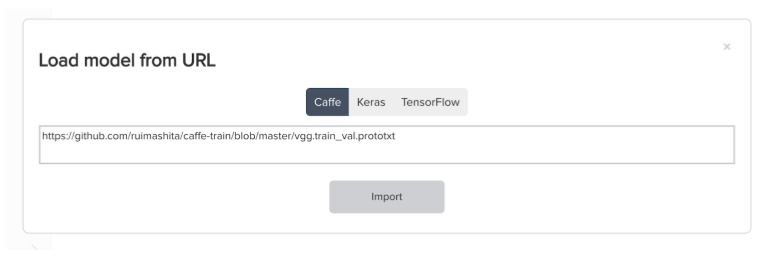}
        (e)
    \end{minipage}
    \begin{minipage}{.75\linewidth}
        \centering
        \includegraphics[width=\linewidth]{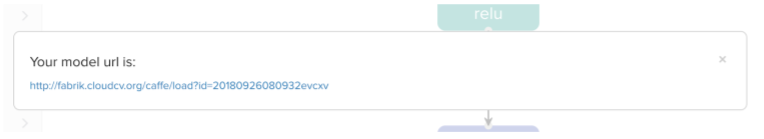}
        (f)
    \end{minipage}
    \caption{Screen-shots of the application depicting the stages in the life-cycle of loading(a,e), visualizing(b), editing(c), saving(d), and sharing(f) a neural network model with Fabrik.}
    \label{fig:walkthrough}
\end{figure*}

\section{Walk-through}
\label{walk-through}

We now provide a complete walk through of the visualizing, editing, and sharing functionalities via a concrete example. Consider that we want to visualize and modify the well known VGG-16 network proposed in ~\citeauthor{vgg-arxiv-2014}. VGG is a 16-layer convolutional neural network proposed for image classification on the ImageNet dataset, that placed second in the ILSVRC classification challenge in 2014. Its architecture is primarily composed of convolutional, pooling, and fully connected layers.

Figure~\ref{fig:walkthrough} walks through this use case: The network is first loaded into the application through the Import option and can be in a format matching any one of the supported frameworks, namely \emph{prototxt} for Caffe, \emph{JSON} for Keras, or \emph{pbtxt} for Tensorflow. Note that we load only the model configuration for visualization, not its parameters. As seen in Figure~\ref{fig:walkthrough}(a), the URL option is selected for import. ~\ref{fig:walkthrough}(e) shows the actual URL, in this case we directly load the VGG-16 model from a prototxt file hosted on GitHub. If the import is successful, the network will be visualized on the canvas as shown in ~\ref{fig:walkthrough}(b). Once the network is loaded, the user can scroll to explore different parts and zoom in/out to see a more local/global view of the network. The user can also hover over any layer to get an overview of layer parameters (~\ref{fig:walkthrough}(b)). If on analyzing the layer parameters, the user wishes to modify a parameter, they can choose to click on that layer. This will launch a parameter pane on the right of the canvas where all layer parameters can be updated as shown in ~\ref{fig:walkthrough}(c). In this case, the number of outputs for the selected convolutional layer are being modified.
Upon editing the network to their satisfaction, the user can choose to export it to any of the supported frameworks or share a link to it with collaborators. Figure~\ref{fig:walkthrough}(d) shows the network being exported back to the Caffe framework from which it was originally loaded. At the bottom-left of the screen we can see the exported output being downloaded as a prototxt file. Lastly, figure~\ref{fig:walkthrough} (f) shows the unique link generated for this model, that can be used to access it in the future, or be shares with others.

\vspace{-10pt}

\section{Additional Use Cases}
\label{useCases}

\subsection{Visualizing Neural Networks}
The core objective of Fabrik is to provide a platform for visualizing neural network models. In this section we elaborate on use cases for these functionalities.

\subsubsection{Importing Existing Networks}
Researchers and deep learning enthusiast often wonder if the code for their neural network model correctly creates the network architecture that they intended. With Fabrik, this will no longer remain a doubt, as they can import their code into Fabrik at the click of a button to visualize its structure and verify correctness. This also enables users to clearly visualize the depth and breadth of their deep learning architecture. Fabrik currently supports several options for import: First, a user can directly upload a model configuration file as \emph{prototxt} (for Caffe), \emph{JSON} (for Keras), or \emph{pbtxt} (for Tensorflow). Note that for TensorFlow and Keras that do not decouple model definition like Caffe does, serializing the model configuration into these formats is easily achieved by functions available in both frameworks. Alternatively, the model configuration can be directly pasted as text into an input form-field. Finally, model configurations available as a URL on GitHub or as a Gist can also be directly imported.



\begin{table*}[t]
\setlength\tabcolsep{10pt}
\centering
\begin{tabular}{@{}p{10cm} c c c@{}} 
\toprule
\textbf{Models}                                 & \textbf{Caffe} & \textbf{Keras} & \textbf{TensorFlow}  \\ 
\midrule
\multicolumn{4}{c}{Recognition}   \\
Inception V3~\cite{szegedy2016rethinking}  & \ding{51}       & \ding{51}       & \ding{51}             \\
Inception V4~\cite{szegedy2017inception}   & \ding{51}       & \ding{51}       & \ding{51}             \\
ResNet 101~\cite{he2016deep}               & \ding{51}       & \ding{51}       & \ding{51}             \\
VGG 16~\cite{vgg-arxiv-2014}               & \ding{51}       & \ding{51}       & \ding{51}             \\
GoogLeNet~\cite{szegedy2015going}          & \ding{51}       & \ding{55}       & \ding{55}             \\
SqueezeNet~\cite{iandola2016squeezenet}    & \ding{51}       & \ding{55}       & \ding{55}             \\
AllCNN~\cite{springenberg2014striving}     & \ding{51}       & \ding{55}       & \ding{55}             \\
DenseNet~\cite{huang2017densely}           & \ding{51}       & \ding{55}       & \ding{55}             \\
AlexNet~\cite{krizhevsky2012imagenet}      & \ding{51}       & \ding{51}       & \ding{51}             \\ 
\midrule
\multicolumn{4}{c}{Detection}                                                                                                                      \\
YoloNet~\cite{redmon2016you}               & \ding{51}       & \ding{51}       & \ding{51}             \\
FCN32 Pascal~\cite{long2015fully}          & \ding{51}       & \ding{55}       & \ding{55}             \\ 
\midrule
\multicolumn{4}{c}{Seq2Seq}                                                                                                                       \\
Pix2Pix~\cite{isola2017image}              & \ding{51}       & \ding{55}       & \ding{55}             \\ 
\midrule
\multicolumn{4}{c}{Visual Question Answering}                                                                                                      \\
VQA~\cite{antol2015vqa}                    & \ding{51}       & \ding{51}       & \ding{51}             \\
\bottomrule
\end{tabular}
\caption{List of tested models on Fabrik}
\label{tab:testedModels}
\end{table*}

\subsubsection{Creating Networks with Drag and Drop}
Fabrik provides a drag and drop neural network creator. The user thus does not need to be familiar with the syntax of particular frameworks, and can simply drag and drop layers from the dashboard onto the canvas to build a network from scratch. Further, they can edit layer parameters and then simply connect the layers to define the computation graph. Once the user is satisfied with the model configuration they can choose to export it to any of the supported frameworks of their choosing. 


\subsubsection{Parameter Calculation}
Most layers in neural networks contain trainable parameters, and the total number of available parameters is correlated with model capacity, as well as the amount of time time training will take. If there are very few trainable parameters, the network might not be able to fully capture the complexity of the data and may end up under-fitting, while too many parameters the network might lead to overfitting. With Fabrik, the total number of parameters for the network being visualized is calculated and displayed in real time at the bottom-left of the canvas. Fabrik updates this count whenever a layer parameters is updated, or when a layer is deleted or added. This can provide an idea of the impact of the addition/removal of a layer on the parameter count, and can serve as a good teaching tool. 

\subsubsection{Exporting Models}
Fabrik supports model export to any of its supported frameworks. Once the user is satisfied with their network configuration, they can select their target framework, and depending on the framework, a prototxt / JSON / pbtxt model configuration file is generated. This model configuration can be loaded into that framework with minimal effort (typically $<5$ lines of code, we provide instructions for inexperienced users), and be further used to train/test that model.

\subsection{Cross Framework Model Conversion}
Deep learning researchers tend to have a preferred framework that have expertise in, or is best suited to their application. They tend to publish code for their research in that framework. When other researchers wish to build upon that work but are unfamiliar with a framework, this proves to be a hindrance. Fabrik can bridge this gap as models can easily be ported to a framework the user is comfortable in. In some cases, this is not possible, particularly when a layer present in one framework does not have an equivalent in other. We overcome this to an extent by using crowd-sourced custom implementation of missing layers. However, with a large number of models, cross framework conversion helps users work in their preferred framework.

\par\noindent
Table~\ref{tab:testedModels} shows the results of model conversion between currently supported frameworks. As seen, certain models cannot be converted to certain frameworks. In the case of GoogLeNet, export to TensorFlow and Keras is not possible due to the LRN layer that only Caffe supports. 

\vspace{-10pt}
\subsection{Real-time Collaboration}
Fabrik provides users with a platform where they can build deep learning models collaboratively in real time. Real time collaboration allows users to share the model they are working on with multiple users with the help of a unique link, which can be used to access and edit the same copy of the model. Further, users can view the history of updates performed on the model in a collaborative session. They can also revert to a specific state of the model at a particular point of time, which assists in debugging. 
Collaborators can also comment on various aspects of the model architecture to provide feedback and make working remotely efficient.

\vspace{-10pt}
\section{Conclusion and Future Work}
\label{conclusion}
In this work we propose Fabrik, a simple GUI for visualizing, creating, editing, and sharing neural network models in the browser. Fabrik provides novel real-time collaboration features that to assist researchers to brainstorming and design models remotely and at scale. 
It provides a framework agnostic visualization scheme based on a novel intermediate representation that can be used to quickly and efficiently visualize model architectures. While of today it supports three popular deep learning frameworks, as future work we plan on adding support for additional popular frameworks, such as PyTorch and Caffe2. Further, we plan to incorporate ONNX as our intermediate representation as a step towards long term interoperability and maintainability. Finally, Fabrik will be available as an online service which provides instant access on any browser supporting device for the purposes of debugging, reproducibility, learning, and collaboration.


\noindent\textbf{Acknowledgements.} We would like to thank Google Summer of Code and the 30+ developers who have contributed code to the Fabrik project. This work was supported in part by NSF, AFRL, DARPA, Siemens, Google, Amazon, ONR YIPs and ONR Grants N00014-16-1-\{2713,2793\}. The views and conclusions contained herein are those of the authors and should not be interpreted as necessarily representing the official policies or endorsements, either expressed or implied, of the U.S. Government, or any sponsor.

\bibliography{mybib}
\bibliographystyle{sysml2019}

\end{document}